\pdfoutput=1
\documentclass[11pt]{article}
\usepackage[table]{xcolor}
\usepackage{authblk}
\makeatletter
\renewcommand\AB@affilsepx{\quad \protect\Affilfont}
\makeatother

\PassOptionsToPackage{numbers, compress}{natbib}

\usepackage[]{acl}

\usepackage{times}
\usepackage{latexsym}

\usepackage[T1]{fontenc}
\usepackage[utf8]{inputenc}
\usepackage{microtype}
\usepackage{listings}
\usepackage{multirow}
\usepackage{hhline}
\usepackage{changepage} % used for adjustwidth (for large tables)
\usepackage{adjustbox}
\usepackage{subcaption} % subfigure is obsolete and doesn't work with listings; using subcaption instead
\usepackage{url}
\usepackage{wrapfig}
\usepackage{booktabs}
\usepackage{siunitx}
\usepackage{amsmath,amsfonts,amssymb}
\usepackage{cleveref}
\usepackage{tabularx}
\usepackage{enumitem}
\usepackage{float}
\usepackage{xspace}
\usepackage{geometry}
\usepackage{longtable}

\definecolor{CBblue}{RGB}{0, 107, 164}
\definecolor{CBorange}{RGB}{255, 128, 14}
\definecolor{CBgray}{RGB}{89, 89, 89}
\definecolor{CByellow}{RGB}{171, 171, 42}
\definecolor{CBpurple}{RGB}{137, 61, 86}
\definecolor{CBgreen}{RGB}{27, 158, 119}
\definecolor{CBpink}{RGB}{215, 48, 39}

\usepackage{tcolorbox}
\usepackage{fontawesome}

\tcbuselibrary{skins}
\newtcolorbox{challsumm}{
  enhanced,
  colframe=red!70!black,
  colback=red!10!white,
  coltitle=white,
  fonttitle=\bfseries,
  sharp corners,
  boxrule=1pt,
  drop fuzzy shadow
}

\newtcolorbox{chall}[1]{
  enhanced,
  colframe=red!70!black,
  colback=red!10!white,
  coltitle=white,
  fonttitle=\bfseries,
  title={\faWarning\hspace{0.5em}\textbf{#1}},
  sharp corners,
  boxrule=1pt,
  drop fuzzy shadow
}

\newtcolorbox{constraint}[1]{
  enhanced,
  colframe=orange!70!black,
  colback=orange!10!white,
  coltitle=white,
  fonttitle=\bfseries,
  title={\faWarning\hspace{0.5em}\textbf{#1}},
  sharp corners,
  boxrule=1pt,
  drop fuzzy shadow
}

\newtcolorbox{constraintsumm}{
  enhanced,
  colframe=orange!70!black,
  colback=orange!10!white,
  coltitle=white,
  fonttitle=\bfseries,
  sharp corners,
  boxrule=1pt,
  drop fuzzy shadow
}

\Crefname{section}{Sec.}{Secs.}
\Crefname{equation}{Eq.}{Eqs.}
\Crefname{figure}{Fig.}{Figs.}
\Crefname{tabular}{Tab.}{Tabs.}

\usepackage{pifont}% http://ctan.org/pkg/pifont

\usepackage{tikz}
\usepackage{fontawesome}

\usepackage[edges]{forest}
\definecolor{hiddendraw}{RGB}{205, 44, 36}
\definecolor{hidden-blue}{RGB}{194,232,247}
\definecolor{hidden-orange}{RGB}{243,202,120}
\definecolor{hidden-yellow}{RGB}{242,244,193}
\definecolor{hidden-red}{RGB}{255,0,0}
\definecolor{hidden-grey}{RGB}{122,122,122}
\usepackage{xcolor}         % colors
\definecolor{mydarkblue}{rgb}{0,0.08,0.55}
\definecolor{DarkRed}{HTML}{780000}
\definecolor{RegRed}{HTML}{C1121F}
\definecolor{CoolBlue}{HTML}{669BBC}
\definecolor{PlotGreen}{RGB}{81,157,62}
\definecolor{PlotOrange}{RGB}{239,133,54}
\definecolor{PlotBlue}{RGB}{59,118,175}
\definecolor{PlotRed}{RGB}{197,57,50}

\tikzstyle{mybox}=[
    rectangle,
    draw=hiddendraw,
    rounded corners,
    text opacity=1,
    minimum height=1.5em,
    minimum width=5em,
    inner sep=2pt,
    align=center,
    fill opacity=.5,
    ]

\hypersetup{
    colorlinks=true,
    linkcolor=blue,
    filecolor=magenta,  
    urlcolor=cyan,
    pdfpagemode=FullScreen,
    }


\title{Review GIDE - Restaurant Review Gastrointestinal Illness Detection and Extraction with Large Language Models}

\author[$\alpha$, $\ast$]{Timothy Laurence}
\author[$\alpha$, $\ast$]{Joshua Harris}
\author[$\alpha$]{Leo Loman}
\author[$\alpha$]{Amy Douglas}
\author[$\alpha$]{\\Yung-Wai Chan}
\author[$\alpha$]{Luke Hounsome}
\author[$\alpha$]{Lesley Larkin}
\author[$\alpha$]{Michael Borowitz}

\affil[$\alpha$]{UK Health Security Agency}

\begin{document}
\maketitle
\thispagestyle{plain}
\pagestyle{plain}
\begin{abstract}

Foodborne gastrointestinal (GI) illness is a common cause of ill health in the UK. However, many cases do not interact with the healthcare system, posing significant challenges for traditional surveillance methods. The growth of publicly available online restaurant reviews and advancements in large language models (LLMs) present potential opportunities to extend disease surveillance by identifying public reports of GI illness.

In this study, we introduce a novel annotation schema, developed with experts in GI illness, applied to the Yelp Open Dataset of reviews. Our annotations extend beyond binary disease detection, to include detailed extraction of information on symptoms and foods. We evaluate the performance of open-weight LLMs across these three tasks: GI illness detection, symptom extraction, and food extraction. We compare this performance to RoBERTa-based classification models fine-tuned specifically for these tasks.

Our results show that using prompt-based approaches, LLMs achieve micro-F1 scores of over 90\% for all three of our tasks. Using prompting alone, we achieve micro-F1 scores that exceed those of smaller fine-tuned models. We further demonstrate the robustness of LLMs in GI illness detection across three bias-focused experiments.

Our results suggest that publicly available review text and LLMs offer substantial potential for public health surveillance of GI illness by enabling highly effective extraction of key information. While LLMs appear to exhibit minimal bias in processing, the inherent limitations of restaurant review data highlight the need for cautious interpretation of results. \newline
\renewcommand*{\thefootnote}{\fnsymbol{footnote}}
\footnotetext[1]{Equal contribution.}
\footnotetext[2]{timothy.laurence@ukhsa.gov.uk}
\renewcommand*{\thefootnote}{\arabic{footnote}}

\end{abstract}
\tableofcontents

\section{Introduction}

Foodborne gastrointestinal (GI) illness is a considerable cause of ill health in the United Kingdom (UK) and around the world. It is estimated that there are 2.4 million cases of foodborne illness in the UK every year, resulting in a cost of £9.1 billion (\$11.8b) \cite{BurdenFoodborneDiseaseUK2018}. However, it is estimated that only 10\% of people experiencing GI illness ever seek medical attention \cite{BurdenFoodborneDiseaseUK2018}. Consequently, at least 90\% of cases of GI illness are typically undetected by public health surveillance systems, which primarily rely on reports of clinical diagnostic samples submitted to frontline laboratories for the identification of causative agents.

The rise of social media means that some people publicly share information about many aspects of their daily lives, including their health status \cite{Vayena2017}. Online restaurant review platforms are a particularly rich potential source of information on whether people contracted foodborne GI illness, as affected diners may want to share their experience with the management of the establishment or other prospective diners. This raises the potential to perform disease surveillance which may give some insight into GI infections not reported to existing surveillance systems.

Since \citet{transformers} developed the Transformer architecture, there have been rapid developments in Natural Language Processing. As these models have scaled in size and performance, it has been possible to use large language models (LLMs) to perform text processing with little or no retraining. Open-weight models like Llama-3.3 and Mistral-Large, and closed-weight models like Claude-3.7 Sonnet, GPT-4o and GPT-o1 have all demonstrated strong performance on a broad range of text processing tasks \cite{dubey2024llama3herdmodels, mistral2024pixtral, anthropic2024claude35sonnet, openai2024learning}. These powerful foundation models, alongside a collection of health specific models, have greatly expanded the possibilities of using LLMs to perform complex text processing tasks in public health surveillance. 

Our contributions within this paper are:

\begin{enumerate} 
[label=\bfseries\arabic*.,leftmargin=*,align=left]
    \item \textit{New annotations:} Generating a set of novel annotations of the Yelp Open Dataset of reviews, in collaboration with experts in the epidemiology and control of foodborne GI illness.
    \item \textit{Novel research questions:} Expanding on previous research focused solely on the presence or absence of reported illness to gain more granular insight into the symptoms and foods reported.
    \item \textit{Evaluating LLMs:} Evaluating the performance of four leading open-weight LLMs in classifying and extracting relevant information in comparison to expert human annotations. 
\end{enumerate} 

\section{Related Work}

\subsection{Classification of gastrointestinal illness in online restaurant reviews}

Previous research analysing GI illness in free-text online reviews has primarily focused on training or fine-tuning smaller models specifically for the binary classification task of determining whether a review or sentence mentions GI illness.

\citet{effland2018discovering} fit a range of classifiers (support vector machine, logistic regression, and random forest) to identify Yelp restaurant reviews that mention a person or multiple people experiencing foodborne GI illness following dining at a given establishment. To train and evaluate the models, over ten thousand reviews were manually annotated by epidemiologists after being collected and filtered. They report 87.0\% F1\footnote{The F1 score is calculated as the harmonic mean of precision (positive predictive value) and recall (sensitivity). The micro-F1 score is calculated by aggregating counts across all labels before computing precision and recall, while the macro-F1 score is calculated by computing the F1 score for each label separately and averaging them, weighting each label equally.} on their sick or not sick task. \citet{karamanolakis2019weakly} then apply hierarchical and non-hierarchical approaches to identify the segments within the reviews that are specifically referring to GI illness. They report a F1 score of 89.6\% on their sick or not classification task. Finally, \citet{liu2020detecting} further build on this dataset in order to fine-tune mBERT and BERT (Bidirectional Encoder Representations from Transformers) language models to identify references to GI illness within multi-lingual reviews. They individually trained mono-lingual BERT models are shown to outperform both the original logistic regression approach~\citep{effland2018discovering} and the jointly trained multi-lingual mBERT model. They find that their best performing BERT model achieves a F1 score of 91.6\%.

\citet{harris2024evaluatinglargelanguagemodels} evaluate the performance of the leading open-weight models on 16 classification and extraction text processing tasks (including a limited initial version of the processing performed on this dataset). They find that Llama-3.3 is the best performing model across most tasks, and that adding examples (or shots) can increase the performance of prompted LLMs at more challenging tasks. 

\subsection{Other health natural language processing}
\label{subsec:other_nlp}
There are a growing number of domain specific models that are optimised for certain domains (with further fine-tuning potential). The first wave of transformer models for clinical text and biomedical sciences included: BioBERT \cite{Lee2019}, SciBERT \cite{beltagy2019scibertpretrainedlanguagemodel}, ClinicalBERT \cite{huang2020clinicalbertmodelingclinicalnotes} from the BERT-family of models. More recently there are open-weight large language models like Meditron \cite{chen2023meditron70bscalingmedicalpretraining}, InternistAI \cite{Griot2024}, which involve fine-tuning Llama-3 and Mistral respectively to perform health text processing. and closed source models including MedPALM-2 \cite{singhal2023expertlevelmedicalquestionanswering} and Med-Gemini \cite{saab2024capabilitiesgeminimodelsmedicine}. These health-specific models are often obtaining State of the Art (SOTA) results at specific health tasks alongside the highest performing foundational models. 

Outside of GI illness, LLMs have been applied in the wider medical domain to classify free text. Many different authors have used LLMs, smaller fine-tuned models and other machine learning models to classify medical text, for instance to provide a diagnosis \cite{saab2024capabilitiesgeminimodelsmedicine, chen2024clinicalbenchllmsbeattraditional, wang2023llmslikegpt4outperform, wang2023drgllama, madadi2023chatgpt, chen2023evaluation, shoham2023cpllm, kefeli2023large}.

Extracting structured information from free text has been a long-term challenge within NLP. Recently, combining general LLMs with in-context learning approaches has been shown to be increasingly effective. \citet{han2024empiricalstudyinformationextraction} set out different forms of text extraction tasks, and how using GPT-4 with prompting only compares to the previous SOTA for 16 dataset task combinations. As well as the more general studies on extraction using LLMs, a number of applications that utilise these approaches have been proposed within the medical domain. With authors again using LLMs, or fine-tuned smaller models to obtain state of the art results in medical text extraction across a range of diseases and types of text \cite{agrawal2022large, doosterlinck2023biodex, bisercic2023interpretable, shyr2023identifying, Hu_2024}. 

\section{Methods}

\subsection{Data}
\label{subsec:data}
To train and evaluate our language models we use a subset of restaurant reviews from the Yelp Open Dataset~\cite{yelpopendataset}. Following a similar approach to \citet{effland2018discovering}, we first filter to only those reviews containing a comprehensive list of possible GI related keywords (<1\% of total reviews). We then randomly sampled c.3,000 reviews to be manually annotated using a protocol designed by UKHSA epidemiologists (see Section \ref{subsec:gi_classification_annotation} for details). 

The resulting dataset is summarised in Table \ref{tab:summary_statistics_gi_detection}. This yields 1,148 reviews (37.4\%) annotated as GI-related and 1,921 (62.6\%) annotated as not-GI-related reviews. We use the term GI-related because the symptoms reported are potentially consistent with GI illness, but there is no microbiological or genomic confirmation of a GI pathogen. We consider this an \emph{adversarial} sample as all reviews contain keywords that could related to GI illness and, as such, it represents a more challenging classification task than a random sample of the original data.
\begin{table}
\centering
\begin{tabular}{lr}
\toprule
                 Statistic &  Value \\
\midrule
             Total Samples &   3069 \\
             Unique Labels &      2 \\
  Mean Text Length (words) &    159 \\
Median Text Length (words) &    120 \\
   Min Text Length (words) &     10 \\
   Max Text Length (words) &    959 \\
              Label: False &   1921 \\
               Label: True &   1148 \\
\bottomrule
\end{tabular}
\caption{Descriptive Statistics for GI Classification}
\label{tab:summary_statistics_gi_detection}
\end{table}

A subset of the reviews labelled as relating to GI are then annotated with the relevant symptoms reported. This is described in Table \ref{tab:summary_statistics_gi_symptoms}. The manually annotated symptoms are disambiguated into five symptom labels: diarrhoea, vomiting, abdominal pain (using a broad definition including different forms of abdominal discomfort), bloody stool and general sickness (which is where the review describes being ill without naming specific symptoms). Bloody stool, which occurs only with a subset of pathogens, was not observed in the manual annotation process. Other symptoms such as headache, fever, and respiratory symptoms were not annotated, because they are not sufficiently specific to GI illness.

\begin{table}
\centering
\begin{tabular}{lr}
\toprule
                 Statistic &  Value \\
\midrule
             Total Samples &    500 \\
             Unique Labels &      5 \\
  Mean Text Length (words) &    129 \\
Median Text Length (words) &     88 \\
   Min Text Length (words) &     10 \\
   Max Text Length (words) &    759 \\
   Label: general sickness &    277 \\
          Label: diarrhoea &    110 \\
           Label: vomiting &     95 \\
     Label: abdominal pain &     55 \\
             Label: nausea &     34 \\
\bottomrule
\end{tabular}
\caption{Descriptive Statistics for Symptom Extraction}
\label{tab:summary_statistics_gi_symptoms}
\end{table}

We then manually annotate these reviews with all the foods mentioned within the text. This is described in Table \ref{tab:summary_statistics_gi_foods}. Foods are very challenging to disambiguate, so a large lookup table of foods based on the FoodEx database was produced and used to disambiguate foods into a list of 27 labels that capture many types of food that could be the source of foodborne GI illness. The data was not annotated to include likely ingredients of food reported, past those that are included in the name of the food. This ingredient mapping is an area where these methods could potentially be extended, but would be subject to considerable uncertainty. 

\begin{table}
\centering
\begin{tabular}{lr}
\toprule
                 Statistic &  Value \\
\midrule
             Total Samples &    500 \\
             Unique Labels &     22 \\
  Mean Text Length (words) &    129 \\
Median Text Length (words) &     88 \\
   Min Text Length (words) &     10 \\
   Max Text Length (words) &    759 \\
              Label: other &    164 \\
            Label: poultry &     88 \\
               Label: meat &     80 \\
          Label: vegetable &     70 \\
               Label: fish &     62 \\
              Label: dairy &     53 \\
              Label: salad &     42 \\
               Label: rice &     35 \\
               Label: beef &     31 \\
          Label: shellfish &     30 \\
              Label: fruit &     22 \\
               Label: eggs &     17 \\
               Label: pork &     14 \\
            Label: lettuce &     11 \\
            Label: sausage &     10 \\
     Label: nuts and seeds &      7 \\
              Label: herbs &      7 \\
         Label: cured meat &      6 \\
         Label: other meat &      4 \\
        Label: mutton/lamb &      4 \\
            Label: berries &      3 \\
 Label: tofu and other soy &      3 \\
\bottomrule
\end{tabular}
\caption{Descriptive Statistics for Food Extraction}
\label{tab:summary_statistics_gi_foods}
\end{table}

For the purpose of optimising prompts the data are partitioned into a small validation set used to optimise prompts (20\%) and a larger test set (80\%). We partition these data to include the larger test  used to estimate overall and label-specific performance most accurately. For the purpose of model fine-tuning all of these datasets are partitioned into training (70\%), validation (10\%) and test sets (20\%). The model is fined-tuned on the training set, hyperparameters (early stopping and learning rate) are  optimised based on the validation set. The test set is used to assess overall and label specific performance. The smaller fine-tuning test set is a subset of the larger prompt-only test set.

Optimised models and prompts are then evaluated on their respective test sets, quantifying accuracy for the best performing prompt / fine-tuned model only. This means the headline test set for the different LLM approaches are not identical, this allows the fine-tuned models sufficient training data, and the best indication of label-level performance of LLMs that are prompted only. The results of the prompt only approaches on the smaller test set is reported in the appendix for completeness.  

\subsection{Approaches to classification and extraction}

\subsubsection{Fine-tuned RoBERTa model}
In an approach similar to \citet{liu2020detecting}, we fine-tune a RoBERTa-large-335m (henceforth referred to as RoBERTa) classifier model to classify whether reviews report GI illness. We fine-tune additional RoBERTa models for extracting symptoms and foods, by reframing these extraction tasks into multi-label classifications. We chose RoBERTa as the base model to fine-tune because it has been demonstrated to perform well on a range of language tasks \cite{liu2019roberta}. Various other authors have also fine-tuned RoBERTa / BERT models to perform text processing tasks in the health-field \cite{kefeli2023large,Lee2019, beltagy2019scibertpretrainedlanguagemodel,
huang2020clinicalbertmodelingclinicalnotes, wang2023drgllama, wei2023chainofthought, Hu_2024, shyr2023identifying}. See Section \ref{subsec:retrain_optim} for detail on our approach to optimising the the fine-tuning.  

\subsubsection{Zero and few-Shot LLM classifier}
As described in Section \ref{subsec:other_nlp}, many LLMs have demonstrated strong performance at health text processing tasks without fine-tuning on the specific task. For each of the models evaluated, we wrote prompts matching their specific prompt template. See Section \ref{subsec:prompt_optim} for details about prompt optimisation on the validation dataset. Examples of correct processing are included in the few-shot prompts. We evaluate a selection of the highest performing open-weight models; we selected a range of sizes of models (8 billion to 123 billion) \cite{dubey2024llama3herdmodels, gemma_2024, cohere_for_ai_2024, mistral2024pixtral}. Due to computational constraints we quantize the largest model Mistral-Large using INT-4 Activation Aware Quantization (AWQ)~\cite{lin2024awqactivationawareweightquantization}, for more information on the deployment approach see \citet{harris2024evaluatinglargelanguagemodels}.

\subsection{Classification of GI illness}
We use a similar approach to \citet{liu2020detecting} focusing on obtaining a boolean (True or False) about whether the review reports GI illness. This is either obtained by the fine-tuned RoBERTa model directly, or by prompting the LLM and disambiguating outputs into a boolean. 

\subsection{Symptom extraction}
Extending on previous work with the Yelp dataset, we also prompt LLMs to extract relevant symptoms from text. The LLM is prompted to extract all symptoms reported by the author as a structured list. The extracted symptoms are disambiguated into the same five symptom labels that the manual annotators applied, with regex in Python. Reviews can be assigned more than one label, except general sickness which can only be applied in the absence of other labels. When fine-tuning a RoBERTa model, each of the five symptoms labels obtained by post-processing manual annotation are used as the targets.

\subsection{Food extraction}
We also evaluate the performance of an LLM at extracting relevant foods. The LLM is prompted to extract all references to food or meals as a structured list. Using the same disambiguation process performed for the manual data (see Section \ref{subsec:data}), these extracted foods are then disambiguated to 27 possible labels, where only 22 of those labels are represented in our dataset. We found that our manually annotated data was very sparse for certain labels, which meant early training experiments were unstable. As such, when fine-tuning the RoBERTa model, we restrict the labels only to 15 that are present in all of the training, validation and test set. We also introduce additional data synthetically generated by GPT-4o to contain reviews and foods; this data is added to the training set only. As a result, the fine-tuned RoBERTa model for food extraction has notable limitations compared to prompt-only methods, due to the constrained label set and reliance on synthetic data generation.  

\section{Results}

\subsection{Overall}

Table \ref{table:micro_f1_scores} shows the results of the prompt optimisation; it shows the overall scores on the validation set for the best performing prompt for each of the model, task, shot combinations. On the GI classification validation set, the best performing model is Mistral-Large achieving 92.3\% micro-F1 with a five-shot prompt. Llama-3-70b is the best performing model at symptom extraction, achieving 92.3\%, again with a five-shot prompt. Command-R-32b performs best at extracting food, where this prompt is zero-shot. Generally adding examples to prompts helps the best performing model, except for the food task where it appears to make all models perform less well. A search was not run over which examples were the most helpful to optimise performance, so potentially optimisation of examples would reverse this effect. Table \ref{table:prompt_fragility_results} shows that prompting the same model with other plausible prompts led to considerable variation in performance, suggesting prompt fragility in obtaining the desired annotations from these models. The models with the most stable results across varying prompts are Llama-3.3-70b and Gemma-2-27b. 

The results of these optimised prompts on the test set is shown in Table \ref{table:micro_macro_f1_scores}. It shows that the Mistral-Large model performs best at the GI classification task with a micro-F1 of 93.4\%. Llama-3.3-70b performs best at symptom extraction with a micro-F1 of 91.5\%. Command-R-32b performs best at the food extraction task with a micro-F1 of 90.6\%. Llama-3.3-70b has the highest mean micro-F1 score across the three tasks. Table \ref{table:micro_macro_f1_scores_reduced} in the appendix confirms similar results are observed on the smaller portion of the test set used to assess the fine-tuned RoBERTa model.  

\begin{table*}[htbt!]
\centering
\resizebox{\textwidth}{!}{%
\begin{tabularx}{\textwidth}{|l|X|X|X|X|X|X|X|X|X|}
\hline
\multirow{2}{*}{Model} & \multicolumn{3}{c|}{Classification} & \multicolumn{3}{c|}{Symptoms} & \multicolumn{3}{c|}{Foods} \\
\cline{2-10}
                       & 0 &  1 &  5 &  0 &  1 &  5 &  0 &  1 &  5 \\
\hline
Mistral-Large       &        0.902 & 0.914 & \textbf{0.923} &       0.836 & 0.918 & 0.918 &    0.873 & 0.845 & 0.852 \\
Llama-3.3-70b &        0.896 & 0.902 & 0.897 &       0.885 & 0.912 & 0.915 &    0.906 & 0.879 & 0.868 \\
Llama-3.1-70b &        0.881 & 0.881 & 0.878 &       0.901 & 0.870 & 0.919 &    0.897 & 0.864 & 0.849 \\
Llama-3-70b   &        0.848 & 0.892 & 0.869 &       0.873 & 0.880 & \textbf{0.923} &    0.878 & 0.858 & 0.832 \\
Command-R-32b     &        0.856 & 0.868 & 0.899 &       0.620 & 0.410 & 0.765 &    \textbf{0.909} & 0.897 & 0.871 \\
Gemma-2-27b         &        0.835 & 0.852 & 0.883 &       0.862 & 0.870 & 0.900 &    0.883 & 0.856 & 0.832 \\
Llama-3.1-8b  &        0.819 & 0.824 & 0.829 &       0.840 & 0.683 & 0.843 &    0.890 & 0.858 & 0.835 \\

\hline
\end{tabularx}%
}
\caption{Statistical performance on the validation set: micro F1 scores for models by task and shot}
\label{table:micro_f1_scores}
\end{table*}

\begin{table*}[htbt!]
\centering
\resizebox{\textwidth}{!}{%
\begin{tabularx}{\textwidth}{|l|X|X|X|X|}
\hline
Model &  Classification &  Symptoms  &  Foods & Mean \\
\hline
Mistral-Large       &              0.106 &             0.298 &          0.048&      0.151 \\
Llama-3.3-70b &              0.129 &             0.064 &          0.079&      0.090 \\
Llama-3.1-70b  &              0.188 &             0.231 &          0.077&      0.165 \\
Llama-3-70b    &              0.223 &             0.371 &          0.488 &      0.361\\
Command-R-32b    &              0.073 &             0.442 &          0.084 &      0.200 \\
Gemma-2-27b        &              0.140 &             0.053 &          0.073 &      0.089 \\
Llama-3.1-8b  &              0.153 &             0.359 &          0.136 &      0.216 \\
\hline
\end{tabularx}%
}
\caption{Difference in micro F1 scores between the highest and lowest performing prompt and shot combination for each task on the validation set}
\label{table:prompt_fragility_results}
\end{table*}
\begin{table*}[htbt!]
\centering
\resizebox{\textwidth}{!}{%
\begin{tabularx}{\textwidth}{|l|X|X|X|X|X|X|X|}
\hline
\multirow{2}{*}{Model} & \multicolumn{2}{c|}{Classification} & \multicolumn{2}{c|}{Symptoms} & \multicolumn{2}{c|}{Foods} &\multicolumn{1}{c|}{All tasks} \\
\cline{2-8}
                       & Micro F1 & Macro F1 & Micro F1 & Macro F1 & Micro F1 & Macro F1 & Micro F1 \\
\hline
Mistral-Large       &        \textbf{0.934} &    \textbf{0.931} &       0.865 &    0.817 &    0.864 &    0.817 &  0.888\\
Llama-3.3-70b &        0.912 &    0.909 &       \textbf{0.915} &    \textbf{0.884} &    0.903 &    0.858 & \textbf{0.910}\\
Llama-3.1-70b &        0.891 &    0.886 &       0.910 &    0.879 &    0.899 &    0.855 & 0.900\\
Llama-3-70b   &        0.904 &    0.900 &       0.884 &    0.853 &    0.871 &    0.780  & 0.887\\
Comm.-R-32b     &        0.911 &    0.906 &       0.774 &    0.737 &    \textbf{0.906} &    \textbf{0.869} & 0.864 \\
Gemma-2-27b         &        0.876 &    0.873 &       0.904 &    0.860 &    0.892 &    0.819 & 0.891\\
Llama-3.1-8b  &        0.816 &    0.815 &       0.862 &    0.689 &    0.888 &    0.853 & 0.855\\
RoBERTa & 0.902 & 0.892 & 0.796 & 0.709 & 0.736* & 0.710* & 0.811\\
\hline
\end{tabularx}%
}
\caption{Statistical performance on the test set: micro and macro F1 scores for models by task. All the models were prompted only except RoBERTa which was fine-tuned. The All tasks micro F1 is the mean of the other micro F1 scores. *Indicates synthetic data was added to the training set}
\label{table:micro_macro_f1_scores}
\end{table*}

Our RoBERTa models fine-tuned to perform these tasks had variable results. For the binary GI classification task the RoBERTa model performs marginally worse than the larger language models that were prompted only, showing that fine-tuning smaller models continues to be near the state of the art for straightforward classification tasks where sufficient annotated data is available. Where we have a smaller sample of annotated examples for symptoms, the fine-tuned model performs markedly worse than prompting only, with a micro-F1 score of 79.6\%. Our initial versions of the food classifier rarely predicted any of the labels; this is because many of the labels occur relatively infrequently with reviews often specifying no foods. Therefore, we added additional synthetically generated examples to the training data, and find this performance improves to a micro-F1 score of 73.6\%.

\subsection{Task specific}

The best performing model for the GI illness classification task is the Mistral-Large model. The granular performance metrics are reported in Table \ref{table:test_detect_label_results}. It shows that the overall F1 score is higher for the Not-GI label, than the GI label. Compare that to the results of the fine-tuned model reported in Table \ref{table:test_detect_ft_label_results}; it shows that the fine-tuned model achieved a worse F1 for both labels. However, the F1 scores for the different methods are within 6\% for both labels. We could have potentially narrowed that gap in performance by annotating additional data to add to the training set for model fine-tuning.

\begin{table*}[htbt!]
\centering
\resizebox{\textwidth}{!}{%
\begin{tabularx}{\textwidth}{|l|X|X|X|X|}
\hline
Label & Precision & Recall & F1-Score & Support \\
\hline
Not-GI      &      0.968 &   0.926 &     0.946 & 1531 \\
GI       &      0.885 &   0.949 &     0.916 &  925 \\
\hline
\end{tabularx}%
}
\caption{GI classification: Performance of Mistral Large prompted-only on the test set. Precision, recall, F1-score, and support for each label (where support is the sample size for that label in the manual annotations)}
\label{table:test_detect_label_results}
\end{table*}
\begin{table*}[htbt!]
\centering
\resizebox{\textwidth}{!}{%
\begin{tabularx}{\textwidth}{|l|X|X|X|X|}
\hline
Label & Precision & Recall & F1-Score & Support \\
\hline
Not-GI     &  0.938     & 0.915  & 0.926    & 410    \\
GI     &  0.837     & 0.878  & 0.857    &  205    \\
\hline
\end{tabularx}%
}
\caption{GI classification report for the fine-tuned RoBERTa model on the test set: precision, recall, F1-score, and support for each label}
\label{table:test_detect_ft_label_results}
\end{table*}

The best performing model for the GI symptoms extraction task is the Llama-3.3-70b model; the granular performance metrics on individual labels are reported in Table \ref{table:test_symptoms_label_results}. It shows strong performance on the general sickness and diarrhoea labels, but much weaker performance for nausea. Whereas when we fine-tune a model on these labels we get the results reported in Table \ref{table:test_ft_symptoms_label_results}. We observe higher F1 scores for nausea (albeit with a very small sample size in the test set) and broadly similar performance for general sickness, but considerably degraded performance for diarrhoea, vomiting and abdominal pain. 

\begin{table*}[htbt!]
\centering
\resizebox{\textwidth}{!}{%
\begin{tabularx}{\textwidth}{|l|X|X|X|X|}
\hline
Label           & Precision & Recall & F1-Score & Support \\
\hline
general sickness &      0.938 &   0.955 &     0.946 &  220 \\
diarrhoea        &      0.940 &   0.868 &     0.903 &   91 \\
vomiting         &      0.888 &   0.934 &     0.910 &   76 \\
abdominal pain   &      0.974 &   0.826 &     0.894 &   46 \\
nausea           &      0.676 &   0.893 &     0.769 &   28 \\
\hline
\end{tabularx}%
}
\caption{GI symptom extraction: Performance of Llama-3-70b prompted-only on the test set. Precision, Recall, F1-Score, and Support for Each Symptom}
\label{table:test_symptoms_label_results}
\end{table*}
\begin{table*}[htbt!]
\centering
\resizebox{\textwidth}{!}{%
\begin{tabularx}{\textwidth}{|l|X|X|X|X|}
\hline
Label                & Precision & Recall & F1-Score & Support \\
\hline
general sickness     &  0.885  &  0.931   &   0.908   & 58    \\
diarrhoea            & 0.750     & 0.632  &  0.686    & 19
\\
vomiting       & 0.474     & 0.818  & 0.600    & 11    
\\
abdominal pain    & 0.429     & 0.500  & 0.462    & 6     
\\
nausea               & 0.800      & 1.000  & 0.889    & 4     \\
\hline
\end{tabularx}%
}
\caption{GI symptom extraction for fine tuned model on the test set: Precision, Recall, F1-Score, and Support for Each Symptom}
\label{table:test_ft_symptoms_label_results}
\end{table*}

The best performing model for the GI food extraction task is the Command-R-32b model. The granular performance metrics are reported in Table \ref{table:test_food_label_results}. The table shows mixed performance for different labels, where many labels are too infrequent to have clear estimates of F1. However, almost all the labels have F1 scores near or above 80\%, and those that do not are all at least 60\%. Whereas for the fine-tuned model, performance is considerably worse, as shown by Table \ref{table:test_retrain_food_label_results}. Unfortunately, we had to remove labels that were not present in all of the train, evaluation and test set. The remaining labels generally have considerably worse performance than the prompt-only model; the F1 scores are generally below 80\%.

\begin{table*}[htbt!]
\centering
\resizebox{\textwidth}{!}{%
\begin{tabularx}{\textwidth}{|l|X|X|X|X|}
\hline
Label                 & Precision & Recall & F1-Score & Support \\
\hline
other              &      0.879 &   0.969 &     0.922 &  128 \\
poultry            &      0.904 &   0.987 &     0.943 &   76 \\
meat               &      0.823 &   0.850 &     0.836 &   60 \\
vegetable          &      0.877 &   0.877 &     0.877 &   57 \\
fish               &      0.936 &   0.936 &     0.936 &   47 \\
dairy              &      0.867 &   0.886 &     0.876 &   44 \\
salad              &      0.966 &   0.933 &     0.949 &   30 \\
rice               &      0.900 &   0.931 &     0.915 &   29 \\
shellfish          &      0.857 &   0.923 &     0.889 &   26 \\
beef               &      0.926 &   1.000 &     0.962 &   25 \\
fruit              &      0.714 &   0.882 &     0.789 &   17 \\
eggs               &      1.000 &   0.929 &     0.963 &   14 \\
pork               &      1.000 &   1.000 &     1.000 &   10 \\
herbs              &      1.000 &   0.714 &     0.833 &    7 \\
sausage            &      1.000 &   1.000 &     1.000 &    6 \\
lettuce            &      0.750 &   0.500 &     0.600 &    6 \\
nuts and seeds     &      0.714 &   1.000 &     0.833 &    5 \\
cured meat         &      1.000 &   1.000 &     1.000 &    4 \\
other meat         &      1.000 &   1.000 &     1.000 &    3 \\
mutton/lamb              &      0.750 &   1.000 &     0.857 &    3 \\
tofu and other soy &      1.000 &   1.000 &     1.000 &    3 \\
berries            &      1.000 &   1.000 &     1.000 &    2 \\
other seafood      &      0.000 &   0.000 &     0.000 &    0 \\
\hline
\end{tabularx}%
}
\caption{Food extraction: Prompt-only model performance on test set. Precision, Recall, F1-Score, and Support for Each Class}
\label{table:test_food_label_results}
\end{table*}
\begin{table*}[htbt!]
\centering
\resizebox{\textwidth}{!}{%
\begin{tabularx}{\textwidth}{|l|X|X|X|X|}
\hline
Label          & Precision & Recall & F1-Score & Support \\
\hline
other          & 0.800     & 0.667  & 0.727    & 30 \\
vegetable      & 0.846     & 0.579  & 0.688    & 19 \\
fish           & 0.900     & 0.600  & 0.720    & 15 \\
poultry        & 0.846     & 0.733  & 0.786    & 15 \\
meat           & 1.000     & 0.500  & 0.667    & 14 \\
dairy          & 0.778     & 0.583  & 0.667    & 12 \\
shellfish      & 0.800     & 0.400  & 0.533    & 10 \\
beef           & 1.000     & 1.000  & 1.000    & 8 \\
rice           & 0.857     & 0.857  & 0.857    & 7 \\
salad          & 1.000     & 1.000  & 1.000    & 6 \\
fruit          & 1.000     & 0.500  & 0.667    & 4 \\
herbs          & 0.000     & 0.000  & 0.000    & 3 \\
cured meat     & 1.000     & 0.500  & 0.667    & 2 \\
pork           & 1.000     & 1.000  & 1.000    & 2 \\
lettuce        & 0.500     & 1.000  & 0.667    & 1 \\
\hline
\end{tabularx}%
}
\caption{Food extraction: Fine-tuned model performance on the test set. Precision, Recall, F1-Score, and Support for Each Class}
\label{table:test_retrain_food_label_results}
\end{table*}

\section{Bias Assessment}
\begin{figure*}[htbt!]
\hspace*{0 cm} % Adjust the value as necessary
    \centering
    \includegraphics[scale=0.6]{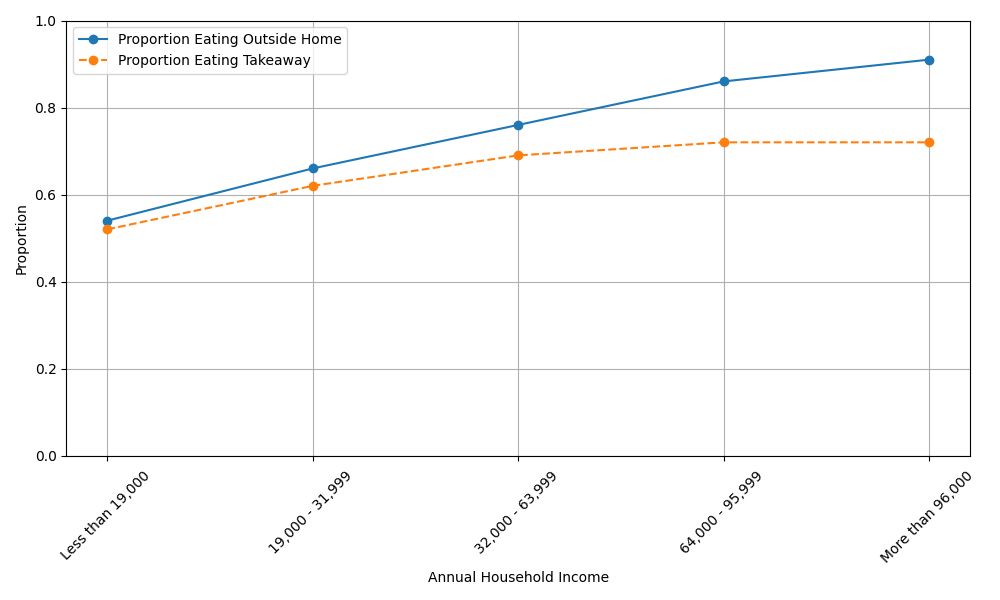} 
    \caption{Data from Food and You 2: Wave Six on the proportion of people eating catered food over the previous four weeks by annual household income \protect\cite{Armstrong2023-px}} 
    \label{fig:eating_out}
\end{figure*}

\begin{figure*}[htbt!]
\hspace*{0 cm} % Adjust the value as necessary
    \centering
    \includegraphics[scale=0.6]{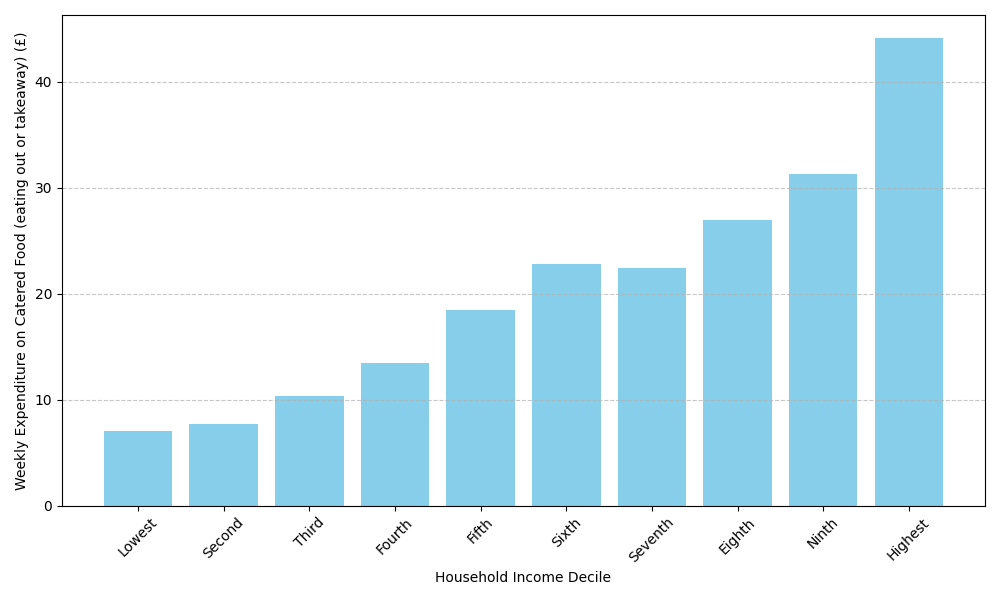} 
    \caption{Data from the ONS on the mean weekly expenditure on catered food by annual household income decile \protect\cite{ONS2024}} 
    \label{fig:expenditure_eating_out}
\end{figure*}

Before conducting these experiments, we anticipated that the main methodological source of bias in this approach would stem from the data. This is because restaurant reviews are a biased sample of possible foodborne GI illness. Figure \ref{fig:eating_out} shows data from the Food Standards Agency on the proportion of individuals consuming catered food in the UK and Figure \ref{fig:expenditure_eating_out} shows the weekly expenditure on catered food by household income decile in the UK. Both figures show a clear income gradient to the consumption of catered food, with higher-income individuals consuming more catered food. The only caveat is that expenditure does not control for price per meal, so the gradient in quantity consumed may be less than the gradient in expenditure. This means that restaurant review surveillance may disproportionately capture GI illnesses among higher-income individuals. It is unclear whether different price points of catered-food are more likely to lead to GI illness or if income is a driver of people posting reviews online. 

As well as the data, we were also concerned about the potential for bias to be introduced by LLMs processing the text. \citet{gallegos2024biasfairnesslargelanguage} set out potential bias concerns from LLMs processing data in a recent literature survey. They set out examples of studies which report degraded (or arbitrary) performance of LLMs at classification and extraction tasks for different demographics \cite{Mozafari2020, mei2023bias}. Where we are using LLMs to generate classifications or extracted text from reviews, we focus our experiments on trying to identify if the text generations of the LLM are invariant under \textit{Social Group Substitution}. \textit{Social Group Substitution} involves changing demographic information and observing the model generations; while expecting invariance is a coarse metric \cite{gallegos2024biasfairnesslargelanguage} it is a reasonable expectation of an unbiased model if the demographic-information is truly arbitrary and the true generations are well-defined. This approach has been used by other authors to evaluate or mitigate bias \cite{rudinger2018genderbiascoreferenceresolution, maudslay2020itsnamemitigatinggender, zhao2024genderbiaslargelanguage}.

We do not process any personally identifiable information as part of this study, as all the reviews are anonymous, without any information like name, address or usernames included. Therefore, if demographic information is used by the LLM, leading to bias between social groups, then it will only be demographic information that can still be inferred from the text of the review. Firstly, the use of certain slang or non-standard spelling may influence the model's performance; this could result from the model's training data lacking sufficient representation of such linguistic variations or because the model infers characteristics of the author based on their writing (for instance, age). Secondly, other demographic information that is likely to be inferred from the text is a person's gender or marital status. As such, we perform three different experiments on versions of the reviews. Firstly, we use Llama-3.3-70b to assess whether spelling and language is traditional or non-traditional (slang use, and unconventional spellings) in reviews. We then stratify reviews by \textit{spelling} to investigate whether there is different performance across these groups. Secondly we filter to reviews which refer exclusively to either men or women. We then map and replace male words to female words (\textit{Social Group Substitution}) and vice-versa; for instance \textit{he} gets mapped to \textit{she}, this yields twice the sample size and allows us to avoid issues of authors writing differently based on gender (as all of the reviews that represent either male or female are now represented in both). And finally we amend terms implying that a couple of diners are married (for instance, \textit{wife}) for the unmarried equivalent (for instance, \textit{girlfriend}); we do this mapping for any of the gender-specific reviews whether they have had the genders remapped or not.

We are not aware of any evidence that would suggest this information (spelling, gender or marital status) \textit{should} affect whether a review is classified as GI or not, and so the performance of the model should be invariant to these factors if the model is performing unbiased text processing in this task. 

Llama-3.3-70b classified 1127 reviews as having \textit{traditional} spelling, and 1329 as \textit{non-traditional}. The performance of our headline model specification hardly varied by \textit{spelling}, with the model achieving 91.2\% micro-F1 where spelling is \textit{traditional} compared to 91.3\% where spelling is \textit{non-traditional}. Under a null-hypothesis that the proportion of reviews classified correctly is the same for both spelling groups, the p-value is 0.93, so is clearly not statistically significant at any reasonable level. Therefore, we do not find evidence that performance is degraded when processing  reviews that contain less \textit{traditional} language. 

\begin{table*}[htbt!]
\centering
\resizebox{\textwidth}{!}{%
\begin{tabularx}{\textwidth}{|l|X|X|X|}
\hline
Spelling &  Micro-F1 & Macro-F1 &  Sample Size \\
\hline
   Standard &     0.912 &     0.903 &         1127 \\
    Non-standard &     0.913 &     0.912 &         1329 \\
\hline
\end{tabularx}%
}
\caption{Micro-F1 and Macro-F1 for the best performing generative prompt stratified by the LLM classification of spelling}
\label{table:bias_detection_spelling}
\end{table*}

The results for our experiments on potential gender bias are shown in Table \ref{table:bias_detection_gender}. They show the micro/macro-F1 for the four different groupings of reviews, where the "synthetic" reviews have had their male terms mapped to female terms or vice versa. It shows there is minimal difference in the overall performance whether the terms used in the reviews relate to men or women, as the micro-F1 are within 0.5\% of each other. Transforming the female unadjusted group of reviews to the male synthetic group leads to no change in accuracy. Transforming the male unadjusted group to female synthetic group leads to a very small increase. Furthermore, statistical analysis of the proportion of correct classifications across these groups does not reveal any significant differences. We conducted both pairwise comparisons of proportions for each of the four groups and comparisons of each group’s proportion to the pooled proportion of correct classifications, with neither approach yielding statistically significant results.

\begin{table*}[htbt!]
\centering
\resizebox{\textwidth}{!}{%
\begin{tabularx}{\textwidth}{|l|X|X|X|}
\hline
Data Grouping & Micro-F1 & Macro-F1 & Sample Size \\
\hline
Male unadjusted   & 0.914 &     0.909 &          429 \\
Male synthetic    & 0.919 &     0.914 &          395 \\
Female unadjusted & 0.919 &     0.914 &          395 \\
Female synthetic  & 0.916 &     0.911 &          429 \\
\hline
\end{tabularx}%
}
\caption{Micro/Macro-F1 for the best performing generative prompt stratified by gender of the terms in that review, and whether the review is amended to be that gender}
\label{table:bias_detection_gender}
\end{table*}

The results for our experiments on potential bias by marital status are shown in Table \ref{table:bias_detection_marital}, to increase sample size we include reviews where the gender terms have been changed in our previous experiment. They show the micro-F1 for the four different groupings of reviews, where the "synthetic" reviews are ones where either the gender or marital status was remapped, compared to the original. Again, it shows there is virtually no difference in the overall performance whether the terms used in the reviews relate to married or unmarried people, as the micro-F1 are within 0.5\% of each other. As with gender, statistical analysis of the proportion of correct classifications across these groups does not reveal any significant differences using either a pairwise or pooled approach.

\begin{table*}[htbt!]
\centering
\resizebox{\textwidth}{!}{%
\begin{tabularx}{\textwidth}{|l|X|X|X|}
\hline
Data Grouping & Micro-F1 & Macro-F1 & Sample Size \\
\hline
Married unadjusted &      0.942 &     0.942 &          190 \\
Married synthetic &      0.944 &     0.944 &          408 \\
Unmarried unadjusted &    0.945 &     0.945 &          109 \\
Unmarried synthetic &     0.947 &     0.947 &          489 \\
\hline
\end{tabularx}%
}
\caption{Micro-F1 and Macro-F1 for the best performing generative prompt stratified by marital status of the terms in that review, and whether the review is amended to be that marital status}
\label{table:bias_detection_marital}
\end{table*}

\section{Discussion}

Overall, we have demonstrated that LLMs can perform relevant text extraction and classification from restaurant reviews to support surveillance of GI illness. We have demonstrated that through prompting only, LLMs can detect GI-related reviews more effectively than smaller-fine-tuned models that represented the previous SOTA. We have also demonstrated that this prompting-only approach can be extended to symptom and foods extraction. This confirms opportunities for GI illness surveillance in online restaurant reviews, to provide another source of intelligence on the burden of foodborne GI illness in England. 

The main potential source of bias in this approach to detecting GI illness is bias in the data itself. Only a subset of all food consumed is catered-food that would potentially generate a review on a website like Yelp. People with higher incomes are more likely to consume catered-food, though it is unclear whether certain types of diners leave reviews more regularly. Imperfect data means the appropriate classification of some reviews is very challenging, even with powerful LLMs performing accurate text processing. 

Having assessed the performance of the highest performing LLM on reviews with variable spelling quality, and with differing arbitrary demographic characteristics, we find no evidence that bias is introduced from the LLM's text processing. 

Another major challenge with review-data itself, is that incubation periods of GI-pathogens can vary from as little as a few hours to several days, meaning people may misattribute their symptoms (that may be caused by a GI pathogen or not) to a specific food service establishment, when it is not actually the source. \citet{Jones2006-nw} also recognise that people may attribute infections to catered food more often than food prepared in other settings, despite evidence non-catered food also leads to foodborne GI illness \cite{Gherman2019NorovirusRisk, Jones2017-vl, Jones2006-nw}. These infections could also have been acquired from a source other than food. This is explored in more detail in Section \ref{subsec:gi_classification_annotation}.

The weakest part of the our text processing is our approach to food disambiguation. The biggest limitation is our focus on named ingredients, rather than likely ingredients of dishes, as most dishes do not spell out their ingredients clearly in the title. Our approach to disambiguating extracted ingredients using a long look up table based on the FoodEx2 database may also lead to some false negatives, where we are unable to match an ingredient to one of our labels. If restaurant reviews were to be used to support disease surveillance, it is likely that these labels could provide very high level indication of foods present in reviews. However, further work is needed to create probabilistic mappings of the names of dishes to potential ingredients. 

Generally, we find prompt only models perform very well. For GI detection, the models perform better than  fine-tuned smaller models that represented the previous SOTA. An advantage of smaller models is still their size and predictability of their outputs. However, for the symptom extraction task and food extraction task, a considerable disadvantage of smaller-tuned models is poor performance on relatively infrequent labels (also known as the long tail problem \cite{han2024empiricalstudyinformationextraction}). This is particularly common in the food extraction task, where we have to add synthetic examples to allow the  model to learn any signal of the correct label through fine-tuning. Even with these synthetic data added, performance is still below that of the larger models with prompting only.   

Where larger models are used with prompting only, manually annotating a subset of data with experts is still a valuable approach. This validation set supports prompt-optimisation, where we find two seemingly similar prompts can have 10ppts difference in model performance at a given task, even for Llama-3-3-70b, the model which we find to be most consistent across prompts. This prompt fragility has been documented by previous work as well \cite{kaddour2023challengesapplicationslargelanguage}. This validation set also ensures that the model is performing the task in a comparable way to that defined by human experts. 

\section{Conclusion}

Our results show that using prompt-based approaches, LLMs achieve over micro-F1 score of over 90\% for all three of our tasks on our test set. Llama-3.3-70b is the model with the best mean micro-F1 score across the three tasks, though Mistral-Large outperforms it on GI classification and Command-R-32B outperforms it on food extraction. Using only prompting, LLMs achieve micro-F1 scores that surpass those of smaller fine-tuned models. We further demonstrate the robustness of Llama-3.3-70b in GI illness detection across three bias-focused experiments.

In conclusion, publicly available review text and LLMs offer substantial potential for public health surveillance of GI illness by enabling highly effective extraction of key information. While LLMs exhibit minimal bias in processing, the inherent limitations of restaurant review data highlight the need for cautious interpretation of results.

\section*{Acknowledgments}
This work was enabled by UKHSA HPC Cloud \& DevOps Technology colleagues developing
and maintaining internal HPC resources. 

\bibliographystyle{acl_natbib}
{\small
\bibliography{refs}
}

\section{Appendix: Protocols}

\subsection{GI classification}
\label{subsec:gi_classification_annotation}

\begin{figure*}[htbt!]
    \centering
    \includegraphics[scale=0.7]{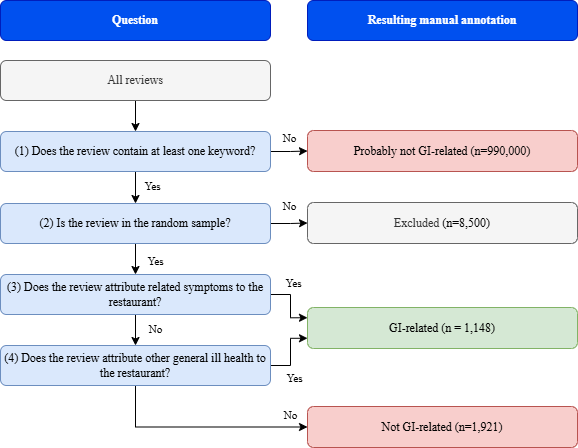}
    \caption{The protocol questions, and resulting manual annotations applied to support the GI classification task}
    \label{fig:protocol_workflow}
\end{figure*}

\Cref{fig:protocol_workflow} shows the steps we undertook to annotate the Yelp dataset with labels to support the GI classification task. Our overall objective was to manually label reviews as either GI-related or not GI-related. We use the term GI-related, as there is no microbiological or genomic confirmation of a GI pathogen. We start with approximately 1,000,000 Yelp Restaurant Reviews. In step (1) we used the following search terms to identify restaurant reviews that are potentially GI-related, an so are in scope to be manually classified: 

["a and e", "a\&e", "accident and emergency", "ambulance", "ambulances", "bad stomach", "barf", "barfed", "barfing", "barfs", "chunder", "chundered", "chundering", "chunders", "diarrhea", "diarrhoea", "doctor", "dodgy stomach", "emergency room", "gastric distress", "healthcare", "hospitalize", "hospitalized", "illness", "nausea", "nauseated", "nauseating", "nauseous", "norovirus", "poisoned", "poisoning", "puke", "puked", "pukes", "puking", "salmonella", "sick", "sicker", "spew", "spewed", "spewing", "spews", "stomach bug", "stomach flu", "stomach issues", "stomach problems", "the er", "the runs", "the shits", "the squits", "threw up", "throw up", "throwing up", "tummy troubles", "unwell", "upset stomach", "upset tummy", "vomit", "vomited", "vomiting", "vomits", "vomitted", "vomitting"]

We required an exact match to include a review, so "sick" would not match with "sickly". We excluded some potentially relevant words because they led to too many false positives: e.g. ill (due to misspelled I'll), anything about faecal matter (because many reviews complain about unclean restrooms) and the term poorly (because it can also describe something not being done well). This yielded approximately 11,500 potentially GI-related reviews. Reviews that did not match any key words are excluded, so false negatives are excluded from analysis along with true negatives, limiting their potential to bias our results. Exclusion of reviews that are GI-related but do not contain one of these words is minor concern as our validation dataset may not be totally representative of the full range of GI-related reviews. 

In step (2) we randomly sampled just over 3,000 reviews because this was considered a large enough sample size. We sampled more in case any needed to be excluded because they were not in English or for other reasons. We ensure in this step that there is no personally identifiable information contained in the reviews before additional processing takes place.

In step (3) we identified whether a review attributes relevant GI symptoms to food prepared by the restaurant, these were classified as GI-related (excluding exceptions). The key symptoms we were looking for were diarrhoea, vomiting, nausea, abdominal pain (or broader discomfort) and bloody stool. The main exception to this step was excluding symptoms that were rationally attributed to non-infection causes, for example: irritable bowel syndrome (IBS), allergies, consuming too much alcohol, pre-existing illness, or unpleasant food due to spiciness, greasiness or other taste issues.

In step (4), we identified whether a review attributes someone becoming generally unwell to the food prepared by the restaurant; these were classified as GI-related (excluding exceptions). We commonly expected to see reports of feeling unwell due to the food or assumed foodborne illness (often referred to as food poisoning), seeking medical care in relation to symptoms which are not specifed, attempts to attribute symptoms which are not described to a GI pathogen. There were a few key exceptions. Firstly, where general feeling of illness was attributed to reasons inconsistent with a GI pathogen, because they refer to other causes (examples of other causes in step (3)). Secondly, if another non-GI symptom was included alongside the statement of illness, these incompatible symptoms would most likely be purely respiratory symptoms.  

This protocol has several key limitations. The most significant is the potential for misattributing symptoms to a specific food or establishment. One major factor contributing to this is the incubation periods of GI pathogens, which often delay the onset of symptoms by several days (see Table \ref{tab:incubation}). This delay increases the likelihood of reviewers mistakenly associating their symptoms with food consumed recently rather than the actual source of infection.

To address this, we explored the feasibility of excluding reviews where the reported timing of symptom onset was implausible (e.g. vomiting during a meal or symptoms appearing three weeks afterwards). However, very few reviews contained sufficient information on timing to make this approach practical. As a result, timing inconsistencies were not included as criteria for exclusion in the protocol.

Beyond timing, other sources of misattribution were also difficult to address. Previous research \citet{Jones2006-nw} highlights a tendency for individuals to disproportionately report / associate infections with catered food rather than food prepared in other settings. This bias persists despite evidence showing that non-catered food can also cause foodborne GI illness \cite{Gherman2019NorovirusRisk, Jones2017-vl,  Jones2006-nw}. Expert epidemiologists specialising in foodborne illness outbreaks confirmed this pattern, observing that individuals often over-attribute their illness to recently consumed food or meals not prepared by themselves.

\begin{table}[h!]
\centering
\small % This command will make the text smaller within the table environment
\begin{tabular}{|p{2cm}|p{5cm}|}
\hline
\textbf{Pathogen} & \textbf{Incubation period} \\
\hline
Norovirus & Usually 12-62 hours (range of 6-84 hours) \\
\hline
Campylobacter & Usually 2-5 days (range of 1-10 days) \\
\hline
Escherichia coli &  Usually 2-4 days for Shiga toxin/verocytotoxin producing Escherichia coli (STEC); other forms reported range from 1 hour to 7 days. Most cases within about 
10-50 hours for Enterotoxigenic (ETEC) and Enteroinvasive (EIEC) or about 8-18 hours for Enteropathogenic (EPEC) and Enteroaggregative (EAEC).\\
\hline
Salmonella & Most commonly 12-48 hours (range of 4-120 hours) \\
\hline
Clostridium perfringens & Usually 8-18 hours (range of 6-24 hours) \\
\hline
Bacterial toxins e.g. Bacillus spp. & 1-5 hours for emetic symptoms 8-16 for diarrhoeal syndrome \\
\hline
\end{tabular}
\caption{Relevant incubation periods for key GI pathogens \citep{UKHSA2020}}
\label{tab:incubation}
\end{table}

\begin{table*}[htbt!]
\centering
\small % This command will make the text smaller within the table environment
\begin{tabular}{|p{5cm}|p{3cm}|p{2cm}|p{5cm}|}
\hline
\textbf{Example} & \textbf{Step} & \textbf{Resulting classification} & \textbf{Classification explained} \\
\hline
We had a wonderful evening, great Italian food! & Step 1 key word search for GI related terms & Not GI-related & This review would have been automatically filtered by the key word search and never assessed \\
\hline
Food was disgusting, I got food poisoning & Step 2 random sample & Excluded & Though this review would be manually annotated as GI-related, some of the reviews were never annotated as we only needed a sample. \\
\hline
Food was horrible, so we didn't eat a lot, the small amount of food that was consumed left us suffering with vomiting and diarrhea. & Step 3 looking for specific symptoms & GI-related & Key symptoms reported were diarrhoea and vomiting, which are GI symptoms of interest, didn't attribute those to any of the exceptions (e.g. allergies) \\
\hline
We all found ourselves with food poisoning. The only authentic thing about this place is the guaranteed authentic  diarrhea and vomit in the aftermath of consuming their food. & Step 3 looking for specific symptoms & GI-related & Key symptoms reported were diarrhoea and vomiting, which are GI symptoms of interest, didn't attribute those to any of the exceptions (e.g. allergies) \\
\hline
A friend ordered the soup, but the waiter didn't say it contained shrimp and he's allergic. Before we even left the restaurant he was vomiting and having vile bowl movements as his face turned completely red with hives. & Step 3 looking for specific symptoms & Not GI-related & Key symptoms reported but attributed to an allergy rather than possible GI pathogen. Symptoms and timing consistent with allergy, but that is not required. \\
\hline
This restaurant gave me and my wife food poisoning & Step 4, where specific symptoms not mentioned, looking for general reports of illness & GI-related & Reported "food poisoning" which we assume to mean GI Foodborne Illness \\
\hline
I got sick after eating at this restaurant & Step 4, where specific symptoms not mentioned, looking for general reports of illness & GI-related & Reported "got sick" which we assume to mean GI Foodborne Illness \\
\hline
I got sick after eating at this restaurant, I had an awful cough and runny nose for a week after eating here & Step 4, where specific GI-symptoms not mentioned, looking for general reports of illness & Not GI-related & Reported "got sick"; however, further context suggests the illness is not GI-related \\
\hline
From all the good reviews, I was looking forward to thin crust and delicious toppings but I would have settled for a decent slice to make it through the eagles game. Instead, I got violently ill with in 20 minutes after finishing one slice & Step 4, where specific GI-symptoms not mentioned, looking for general reports of illness & GI-related & Reported "getting violently ill"; however, it was only after 20 minutes which does not appear consistent with any common GI pathogens. However, there is rarely enough context to apply this exception consistently. \\
\hline
\end{tabular}
\caption{Examples of manual classification of reviews based on GI-related text}
\label{tab:classification_examples}
\end{table*}

\subsection{Symptom extraction}

In order to manually annotate data with relevant symptoms, we have to decide which symptoms are of interest and which are not. We reviewed classifications from ICD-10 as well as internal UKHSA documentation on GI pathogens and lists of symptoms from other public health agencies. We focus on assigning the following five labels: \textit{diarrhoea}, \textit{bloody stool}, \textit{vomiting}, \textit{abdominal pain} and \textit{nausea}. 

There are other relevant symptoms that are associated with several GI pathogens, but are non-specific to GI, so are also likely to be something else: Fever, Myalgia (muscle aches), Headache, Fatigue, and Loss of appetite. We did not include bloating or flatulence specifically (we included this in abdominal pain), because often people are not specific enough to discern these symptoms. 

We also reviewed and disregarded the following other symptoms that are much more likely to be associated with another foodborne contaminant or non-GI infection: Allergic reactions, Chills and sweats, Cough, Sore throat, Shortness of breath, Nasal congestion, Stiffness (neck or otherwise), Genito-urinary symptoms, Rash, itch or other skin complaint, Other pain that is not abdominal, Swollen lymph nodes, Eye complaints, Arthralgia, Confusion or altered mental state, Change in vision.

The steps to manually annotate the data for symptoms are: Step (1) filter to reviews that have been manually annotated as GI-related.
Step (2) randomly sample 500 reviews. 
Step (3) manually annotate the review with appropriate labels where a review can have up to five labels. Reviews with no specific symptoms reported are labelled general sickness. 
Step (4) apply a post processing function to ensure any extracted symptoms are disambiguated to the exact wording of the pre-defined labels to allow for consistent follow on analyses (see Table \ref{tab:symptom_disambig}.

\begin{table}[h!]
\centering
\small % This command will make the text smaller within the table environment
\begin{tabular}{|p{2cm}|p{5cm}|}
\hline
\textbf{Symptom} & \textbf{(Parts of) words to match} \\ \hline
Vomiting & vom, throw, threw, hurl, barf, chunder, puk \\ \hline
Diarrhoea & diar, run, shits, toilet, upset tummy, upset stomach, stomach issues, stomach problems \\ \hline
Abdominal Pain & abdomin, stomach ache, stomach pain, stomach cramps, tummy ache, belly ache, tummy pain, belly pain \\ \hline
Bloody Stool & bloody stool, bloody poop, bloody poo, red stool, red poop, red poo \\ \hline
Nausea & naus, queas \\ \hline
General Sickness & This code is used if nothing matches \\ \hline
\hline
\end{tabular}
\caption{Key words (or parts of words) to disambiguate between symptoms}
\label{tab:symptom_disambig}
\end{table}

\begin{table*}[htbt!]
\centering
\small % This command will make the text smaller within the table environment
\begin{tabular}{|p{5cm}|p{3cm}|p{2cm}|p{5cm}|}
\hline
\textbf{Example} & \textbf{Step} & \textbf{Result of the step} & \textbf{Step result explained} \\ \hline
I had a lovely meal & Step 1 filter to GI-related only & Excluded & This would have been filtered out as non-GI related \\ \hline
Food was disgusting, I got food poisoning & Step 2 random sample & Excluded & Though this review clearly would be manually annotated as GI-related, some of reviews were never annotated as we only needed a sample of 500. \\ \hline
Food was horrible, so we didn't eat a lot, the small amount of food that was consumed left us suffering with vomiting and diarrhea. & Step 3 extraction of symptoms & vomiting, diarrhea & Mentions vomiting and diarrhea so they are transcribed by manual annotator \\ \hline
We all found ourselves with food poisoning. Guaranteed diarrhea and puking in the aftermath of consuming their food. & Step 3 extraction of symptoms & food poisoning, diarrhea, puking & Transcribe all mentions to ill health and symptoms \\ \hline
I got sick after eating at this restaurant & Step 3 extraction of symptoms & got sick & Transcribe all mentions to ill health \\ \hline
Food was horrible, so we didn't eat a lot, the small amount of food that was consumed left us suffering with vomiting and diarrhea. & Step 4 disambiguation of symptoms & vomiting, diarrhoea & Leave vomiting unchanged because "vom" is a key term for vomiting, "diarr" in diarrhea is a key term for diarrhoea so we disambiguate diarrhea to diarrh\textbf{o}ea and do not mix British and American spellings in later analysis \\ \hline
We all found ourselves with food poisoning. Guaranteed diarrhea and puking in the aftermath of consuming their food. & Step 4 disambiguation of symptoms & vomiting, diarrhoea & Puking disambiguates to vomiting because "puk" is a relevant key term, "diarr" in diarrhea is a key term for  diarrhoea so we do not mix British and American spellings in later analysis \\ \hline
I got sick after eating at this restaurant & Step 4 disambiguation of symptoms & general sickness & Got sick doesn't match any of the symptom specific parts of words, so general sickness is the term label assigned. \\ \hline
\end{tabular}
\caption{Steps and results of filtering and extracting symptoms from reviews.}
\label{table:symptom_examples}
\end{table*}

\subsection{Food extraction}

Processing text related to food in order to inform public health action is very challenging. The main food-text related challenges are: (1) disambiguation, the same dish can potentially be described in many variations, (2) variable ingredients, a dish mentioned in a review could contain many different standard or non-standard ingredients and (3) contamination of dishes with unintended ingredients.  

We decided that linking named dishes to ingredients is beyond the scope of this paper, as the type of statistical analysis of recipes to ascertain likely ingredients in dishes is a large distinct natural language processing project in its own right. Contamination is also just an inevitable limitation, as individuals can never report food they do not know they consumed. To simplify the disambiguation task, we adopt broad food labels (set out in Table \ref{tab:food_labels}) aiming to capture the key categories of foods that may cause foodborne GI illness. While broad labels, they would be specific enough to narrow investigations towards high-level categories of food. However, building on this to incorporate more granular coding and likely ingredients could be an important area of research to support usability in practice. 

\begin{figure*}[htbt!]
\hspace*{0 cm} % Adjust the value as necessary
    \centering
    \includegraphics[scale=0.5]{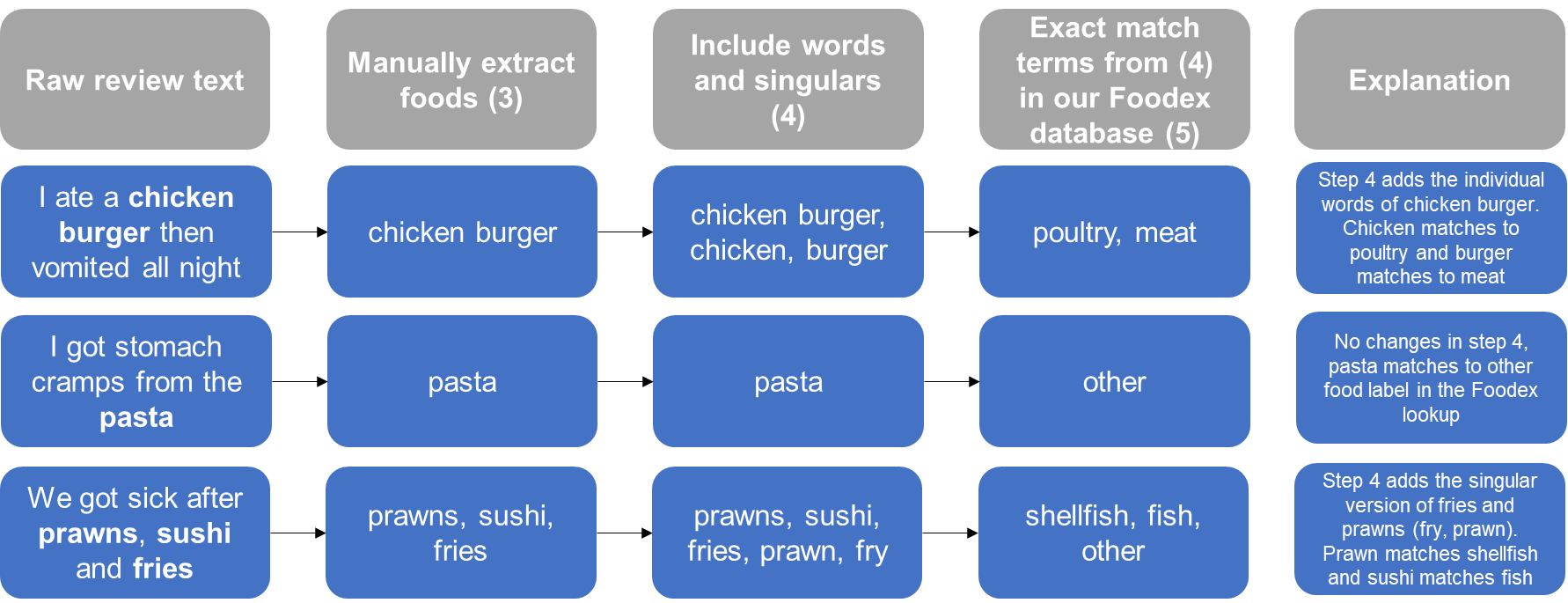} 
    \caption{Examples of how raw text on food is processed} 
    \label{fig:annotation_food_examples}
\end{figure*}

To support manual labelling, we use a version of the FoodEx2 database \cite{efsa2015}. We manually assigned our labels to all approximately 10,000 food items in FoodEx. Food items in FoodEx are generally the ingredient level, as opposed to other levels of abstraction like recipes, we also added in some common terms like sushi, burgers etc. to match with relevant FoodEx codes. We create several different versions of the FoodEx descriptions: we remove all clarifiers so \textit{chicken (fresh meat)} -> \textit{chicken} where the clarifier in brackets is removed to support more exact matches. We create a singular version of the description as well. These altered versions are included in our lookup table of key words to support automated tagging of meals to FoodEx codes. 
The overall process for manual annotation for foods is: Step (1) filter to reviews where GI illness was reported, this means we will have more relevant reviews for identifying higher-risk food stuffs. Step (2) randomly sample 500 reviews to manually annotate. Step (3) manually extract spans of text that report certain food items. Step (4) add any words within the extracted terms to the extracted terms, also add the singular version of any words. Step (5) search all of the terms from step (4) for exact matches in the FoodEx lookup table. This means we are matching foods to labels based on named ingredients only, see \ref{fig:annotation_food_examples} for examples of how text about food is extracted and disambiguated to labels.

As a result of this we have both experimental FoodEx codes (not reported here) but also public health relevant labels. 

\begin{table*}[htbt!]
\centering
\small % This command will make the text smaller within the table environment
\begin{tabular}{p{3cm}p{3cm}p{8cm}}
    \toprule
    \textbf{Broad group} & \textbf{Our label} & \textbf{Reason} \\
    \midrule
    Meat & meat & Risk of transmission of pathogens like Salmonella spp. and Shiga toxin-producing E. coli, enterotoxin producing bacteria such as Clostridium perfringens and Staphylococcus aureus. \\
    Meat & beef & Risk of transmission of Salmonella spp. and Shiga toxin-producing E. coli. \\
    Meat & other meat & Risk of transmission of pathogens like Salmonella spp. and Shiga toxin-producing E. coli. \\
    Meat & pork & Risk of transmission of Salmonella spp., Yersinia enterocolitica. and Hepatitis E. \\
    Meat & mutton/lamb & Risk of transmission of Salmonella spp. and Shiga toxin-producing E. coli.\\
    Meat & poultry & Risk of transmission of Salmonella spp. and Campylobacter spp.  \\
    Meat & cured meat & Risk of transmission of Salmonella spp. and Listeria monocytogenes. \\
    Meat & sausage & Same as risks of pork above. \\
    Meat & other processed meat & Risk of various pathogens listed under other meats and Listeria monocytogenes. \\
    Seafood & seafood & Risk of various pathogens listed under other seafood labels \\
    Seafood & fish & Risk of ingestion of marine biotoxins, histamine and scrombotoxin, Bacillus spp. and Clostridium botulinum.\\
    Seafood & shellfish & Risk of transmission of Vibrio species and norovirus. \\
    Seafood & other seafood & Similar risks to those for fish and shellfish \\
    Other animal products & dairy & Risk of transmission of Salmonella spp., Campylobacter spp., Shiga toxin-producing E. coli and Bacillus species. \\
    Other animal products & eggs & Risk of transmission of Salmonella spp. \\
    Plants or grains & rice & Risk of Bacillus cereus and other enterotoxin producing bacteria. \\
    Plants or grains & vegetable & Risk of transmission of Shiga toxin-producing E. coli, Listeria monocytogenes, Cryptosporidium spp. \\
    Plants or grains & lettuce & Risk of transmission of Shiga toxin-producing E. coli and Cryptosporidium. \\
    Plants or grains & salad & Risk of transmission of Shiga toxin-producing E. coli, Salmonella spp., Shigella spp. and Cryptosporidium. \\
    Plants or grains & sprouts & Risk of transmission of Salmonella spp. and  Shiga toxin-producing E. coli.\\
    Plants or grains & funghi & Risk of natural toxins or poisoning from misidentified species and Listeria monocytogenes. \\
    Plants or grains & herbs & Risk of transmission of Shiga toxin-producing E. coli, Salmonella spp., Shigella spp. and Cryptosporidium. \\
    Plants or grains & nuts and seeds & Risk of transmission of Salmonella spp. \\
    Plants or grains & tofu and other soy & Risk of transmission of Salmonella spp. \\
    Plants or grains & fruit & Risk of transmission of Salmonella spp., Shiga toxin-producing E. coli and Listeria monocytogenes \\
    Plants or grains & berries & Risk of transmission of Shiga toxin-producing E. coli, norovirus and Hepatitis A. \\
    Other & other & Included for completeness, no specific risks \\
    \bottomrule
\end{tabular}
\caption{Food safety rationale for the choice of labels \cite{UKHSA2020, efsa_2024_food, Hawker2018, McLauchlin2007}}
\label{tab:food_labels}
\end{table*}

\section{Appendix: Model Optimisations}

\subsection{Fine-tuning optimisation}
\label{subsec:retrain_optim}
The best performing implementation involves retraining both the final transformer layer and the previously untrained final classification head. We optimise the learning rate using a grid search over feasible candidates, where the model with the highest micro-F1 score on the validation set is chosen. The optimal learning rates are 1e-3 for GI Detection, 3.3e-3 for GI symptoms, and 1e-3 for Foods. The other hyperparameters for this implementation are fixed as a training batch size of 128 and a maximum of 30 epochs of training on the entire training set, where the retraining can stop early when loss on the validation set does not fall over 3 consecutive epochs.

\subsection{Prompt optimisation}
\label{subsec:prompt_optim}

Prompts are optimised using the validation dataset only, this optimisation involves iterative redrafting to ensure the model is generating information in the expected format (e.g. generating a single word if that is requested). We ran a search over different prompt choosing prompts with the highest micro-F1 on our validation dataset. Examples of correct processing are included in the few-shot prompts. These examples are included based on their brevity and breadth; they are standardised across models. 

\section{Appendix: Additional Results}

Table \ref{table:micro_macro_f1_scores_reduced} shows the results of the models on the reduced test set, this test set is reduced to allow for more training data for the fine-tuned model. These results offer greater consistency between the prompt-only approaches and the fine-tuned approach, at the expense of reduced sample size for the prompt-only approaches, where prompt optimisation was performed on a small validation set only. 
\begin{table*}[htbt!]
\centering
\resizebox{\textwidth}{!}{%
\begin{tabularx}{\textwidth}{|l|X|X|X|X|X|X|}
\hline
\multirow{2}{*}{Model} & \multicolumn{2}{c|}{Classification} & \multicolumn{2}{c|}{Symptoms} & \multicolumn{2}{c|}{Foods} \\
\cline{2-7}
                       & Micro F1 & Macro F1 & Micro F1 & Macro F1 & Micro F1 & Macro F1 \\
\hline
Mistral-Large       &        \textbf{0.927 }&    \textbf{0.922} &       0.883 &    0.826 &    0.847 &    0.783 \\
Llama-3.3-70b &        0.907 &    0.901 &       0.922 &    \textbf{0.894} &    0.887 &    0.831 \\
Llama-3.1-70b &        0.880 &    0.872 &       \textbf{0.932} &    0.890 &    0.885 &    0.843 \\
Llama-3-70b   &        0.909 &    0.902 &       0.912 &    0.863 &    0.860 &    0.807 \\
Command-R-32b     &        0.902 &    0.894 &       0.825 &    0.777 &    \textbf{0.902} &    \textbf{0.859} \\
Gemma-2-27b         &        0.875 &    0.869 &       0.917 &    0.848 &    0.860 &    0.756 \\
Llama-3.1-8b   &        0.810 &    0.806 &       0.869 &    0.798 &    0.878 &    0.846 \\
RoBERTa fine-tuned & 0.902 & 0.892 & 0.796 & 0.709 & 0.736* & 0.710* \\
\hline
\end{tabularx}%
}
\caption{Statistical performance on the reduced test set: micro and macro F1 scores for models by task. *Means synthetic data was added to the training set}
\label{table:micro_macro_f1_scores_reduced}
\end{table*}

\end{document}